\pdfoutput=1

\documentclass[11pt]{article}

\usepackage{ACL2023}

\usepackage{times}
\usepackage{latexsym}
\usepackage{hyperref}
\usepackage{tabularx}
\usepackage{graphicx}

\usepackage[T1]{fontenc}

\usepackage[utf8]{inputenc}

\usepackage{microtype}

\usepackage{inconsolata}

\title{Overview of the 2023 ICON Shared Task on Gendered Abuse Detection in Indic Languages}

\author{Aatman Vaidya \\ Tattle Civic Tech \\ aatman@tattle.co.in 
        \And
        Arnav Arora \\ University of Copenhagen \\  aar@di.ku.dk
         \AND  
        Aditya Joshi \\ University of New South Wales \\ aditya.joshi@unsw.edu.au
        \And 
        Tarunima Prabhakar \\ Tattle Civic Tech \\ tarunima@tattle.co.in}

\begin{document}
\maketitle
\section*{Abstract}
This paper reports the findings of the ICON 2023 on Gendered Abuse Detection in Indic Languages. The shared task deals with the detection of gendered abuse in online text. The shared task was conducted as a part of ICON 2023, based on a novel dataset in Hindi, Tamil and the Indian dialect of English. The participants were given three sub-tasks with the train dataset consisting of approximately 6500 posts sourced from Twitter. For the test set, approximately 1200 posts were provided. The shared task received a total of 9 registrations. The best F-1 scores are 0.616 for subtask 1, 0.572 for subtask 2 and, 0.616 and 0.582 for subtask 3.

\textbf{\textit{The paper contains examples of hateful content owing to its topic.}}

\section{Introduction}
\label{sec:intro}
Online gender-based violence is a growing challenge that compounds existing social and economic vulnerabilities. It can cause people to recede from online spaces impacting their political and economic opportunity. At its worst, it can lead to loss of life. Hate speech and gender abuse online can lead to real-world violence \cite{mirchandani2018digital, byman2021hateful, kumar2018aggression}. While there is a need for automated approaches to detect gendered abuse, there is a lack of Indic language datasets that enable such approaches for Indian language content.

At ICON 2023, we conducted a shared task\footnote{\url{https://sites.google.com/view/icon2023-tattle-sharedtask}} led by Tattle Civic Tech\footnote{\url{https://tattle.co.in/}}, based on a novel dataset on gendered abuse in Hindi, Tamil and Indian English. 

The dataset \cite{arora2023uli} provided is created under the Uli\footnote{\url{https://uli.tattle.co.in/}} project. Uli is a browser plugin that de-normalizes the everyday violence that people of marginalised genders experience online in India. Uli also provides tools for relief and collective response. One of its features is to allow users to moderate instances of online gender-based violence in Indian languages. Through focused grouped with over 30+ activists and researchers who have been either at the receiving end of violence or have been involved in making social media more accessible, the pilot team of Uli learnt how online gender-based violence is experienced by different users online. Online harassment leads to people at the receiving end of abuse facing consistent fatigue, panic, and anxiety. Fatigue resulting from hate speech was the most prominent affective response that was noted. 

The dataset contains posts tagged with the following three labels:
\begin{itemize}
\item \emph{Label 1:} Is the post a gendered abuse when directed at a person of marginalized gender?
\item \emph{Label 2:} Is the post a gendered abuse when it is not directed at a person of marginalized gender?
\item \emph{Label 3:} Does this post contain explicit/ aggressive language?
\end{itemize}

These are not mutually exclusive labels, but rather an attempt to capture different ways of understanding gendered abuse. 

The values for each label could be the following, 
``1" indicates the annotator believes the post (tweet) matches the label. 
``0" indicates the annotator does not believe the post (tweet) does not match the label. 
"NL" means the post was assigned to the annotator but not annotated.
"NaN" indicates the post was not assigned to the annotator.

Below are some examples of a post labelled as $1$ or $0$ for each label. Examples of posts annotated as \textbf{1} for all the labels
\begin{itemize}
    \item \textit{Label 1}: \#WomenAreTrash  they must be arrested and throw away the key
    \item \textit{Label 2}: \#Julie stupid girls....horrible girls...420's...culprite...wat a rude behaviour....dnt show u r face...in public.
    \item \textit{Label 3}: .. mmmh these bitches gay. Good for them, good for them
\end{itemize}

Examples of posts annotated as \textbf{0} for all the labels
\begin{itemize}
    \item \textit{Label 1}: Hello mem how are you \#JacquelineFernandez
    \item \textit{Label 2}: ...but all superheroes can't be woman ! More power to u
    \item \textit{Label 3}: ""cannot even burn the effigy"" LMAO
\end{itemize}

\section{Task Description}
The shared task was to develop gendered abuse detection models based on the three labels in the training dataset. This involves the following three subtasks:
\begin{itemize}

\item \textbf{Subtask 1}: Build a classifier using the provided dataset \textit{only} to detect gendered abuse (label 1)
\item \textbf{Subtask 2}: Use transfer learning from other open datasets for hate-speech and toxic language detection in Indic languages to build a classifier to detect gendered abuse (label 1)
\item \textbf{Subtask 3}: Build a multi-task classifier that jointly predicts both gendered abuse (label 1) and explicit language (label 3)
\end{itemize}

\subsection{Task Setup and Schedule}
The shared task was hosted on Kaggle as a Kaggle competition\footnote{\url{https://www.kaggle.com/competitions/gendered-abuse-detection-shared-task}}. Participants were allowed to take part in all the 3 subtasks. If they chose to participate in a subtask, they were required to submit the predictions of the classifier for all three languages. The competition was open to the public, but participants needed to register to qualify for the shared task. Registered participants could access the training and testing dataset through the platform itself. 

Kaggle competitions include an automated evaluation feature that requires the hosts to upload a solution file containing the ground truth values for the test data and the platform automatically calculates the error score for a submission made by the participants. This was one of the major limitations for us as a single Kaggle competition could only facilitate one sub-task. For sub-task three where the results have to be evaluated against 2 test sets, we could not conduct this sub-task on the kaggle. 

The participants were given 3 weeks to develop, experiment and build their classifiers. After 3 weeks, the test set was released, after which the participants had 4 days to test, evaluate and upload their systems. The participants then had to submit a short paper outlining their methodology. The entire timeline and schedule of the shared task is given in Table \ref{tab:task_schedule}.

\begin{table}[h]
\centering
\begin{tabular}{ll}
\hline
\textbf{Event}        & \textbf{Date}      \\ \hline
Training Set Release  & 15th November 2023 \\ \hline
Test Set Release      & 6th December 2023  \\ \hline
Submissions Due       & 9th December 2023  \\ \hline
Results Declared      & 10th December 2023 \\ \hline
Paper Submissions Due & 12th December 2023 \\ \hline
\end{tabular}
\caption{Timeline of the Shared Task}
\label{tab:task_schedule}
\end{table}

In the testing phase, participants were allowed to make submissions upto 5 times a day and their best run was included in the final leaderboard. The leaderboard was also public.

\section{Related Work}
Past work has primarily been done around creating datasets and classifiers for abuse detection. In this section, we look at some of the relevant work. Studies have looked at trolling \cite{mojica2016modeling, kumar2014accurately}, misogyny \cite{frenda2019online}, offensive language \cite{zampieri2019predicting}, cyberbullying \cite{dadvar2013improving} etc. These terms have been used overlapping categories \cite{waseem2017understanding}.

\cite{mandl2019overview, mandl2020overview} proposed dataset for Hate Speech in Hindi language consisting of 5K and 6K posts respectively, \cite{saroj2020indian, velankar2021hate} also contributed datasets for Hindi. \cite{chakravarthi2021findings, bhattacharya2020developing, romim2021hate, gupta2022multilingual} are some other datasets for Indic languages.  

This shared task is one of many shared tasks that
are being organised in similar area. Some other shared tasks include \cite{kumar2020evaluating, kumar2021comma, zampieri2019predicting, zampieri2019semeval, mandl2019overview, mandl2020overview, mandl2021overview, modha2021overview, chakravarthi2021findings, kumar2018proceedings}. \cite{zampieri2019semeval} started with a subtasks model for the shared task which was adopted by other shared tasks as well. The HASOC shared task \cite{mandl2021overview} is a well-known
series of competitions around Hate Speech and Offensive Content Identification detection in English, Hindi, and Marathi. The Dravidian language shared task \cite{chakravarthi2021findings} looked at offensive language detection in Tamil, Malayalam and Kannada. \cite{kumar2018proceedings} shared task looked at trolling, cyberbullying, flaming. Broadly, previous shared tasks look at different aspects of hate speech such as trolling, offensive language, aggression etc. This shared task specifically looks at detecting online gender-based abuse. The dataset provided is also annotated with questions specifically around online gender-based violence. This opens up new directions for future research on detecting abuse in Indic languages.

\section{Dataset}
The dataset \footnote{\url{https://github.com/tattle-made/uli_dataset}} \cite{arora2023uli} contains a total of 7638 posts in English, 7714 posts in Hindi, and 7914 posts in Tamil annotated for 3 labels i.e. each of the 7638 posts in English, 7714 posts in Hindi, and 7914 posts in Tamil have annotations for three labels. Each label is explained in section \ref{sec:intro} of the paper. The subtasks for the shared task were created around label 1 and label 3. 

\begin{table}[h]
\centering
\begin{tabular}{|c|cc|}
\hline
\textbf{Language} & \multicolumn{2}{c|}{\textbf{Split}}                 \\ \hline
                  & \multicolumn{1}{c|}{\textbf{Train}} & \textbf{Test} \\ \hline
English           & \multicolumn{1}{c|}{6531}           & 1107          \\ \hline
Hindi             & \multicolumn{1}{c|}{6197}           & 1516          \\ \hline
Tamil             & \multicolumn{1}{c|}{6779}           & 1135          \\ \hline
\end{tabular}
\caption{Dataset Statistics}
\label{tab:dataset}
\end{table}

This dataset was annotated by eighteen activists and researchers who have faced or studied gendered abuse. The activists and researchers represent a range of socio-cultural as well as geographical backgrounds. During the process of annotation, an annotator could skip a question (label) given to them. 
This dataset, inspired by values of feminist technologies such as inclusion, intersectionality, and care, is an attempt at participatory models of machine learning development.

The training and testing set consists of posts (tweets) sourced from Twitter. All the posts in the dataset have at least one annotation present for each label. The training set has at least one annotation present for each label, there are few posts in the training set with more than one annotation. The posts in the test set contained three annotations for each label.

\section{Participating Teams}
A total of 9 teams registered for the shared task. Each team could choose which subtask(s) they wished to attempt. Once a subtask was chosen, participants were required to attempt it for all three languages. Finally, 2 teams submitted their systems. The teams had to submit a paper outlining the methodology, models, and experiments. In this section, we provide a summary of each team's system. 

Team \textbf{CNLP-NITS-PP} made a submission for all the three subtasks. The team used an ensemble approach built upon a Convolutional Neural Network (CNN) and Bidirectional Long
Short-Term Memory (BiLSTM) architecture for all the three subtasks. For the initial input layers, they used pretrained
GloVe and FastText embeddings of 300-dimensional dense vectors, with the sequence length capped at 100 words. For subtask 2, the team utilized the Multilingual Abusive Comment Detection (MACD) \cite{gupta2022multilingual} dataset for Hindi and Tamil, along with the MULTILATE\footnote{\url{https://github.com/advaithavetagiri/MULTILATE}} dataset for English, as external open datasets for transfer learning, in addition to the provided dataset. The models were trained using the Adam Optimiser and Categorical Crossentropy as the loss function.

Team \textbf{SCalAR} made a submission for subtask 1. The team used BiLSTM architecture. They used fastText word embedding for the initial input layers. These embeddings
were fine-tuned during the training. They employed the
Adam optimizer for efficient gradient-based optimization and categorical cross-entropy loss as a loss function.
\section{Results}
The systems were evaluated based on F-1 score error metric. The teams' system results were considered in two ways: their F-1 score, reflecting their rank on the leaderboard, and their paper submission describing their methodology. The results of both the teams. The results are listed in Table \ref{tab:results}.

\begin{table}[h]
\centering
\resizebox{\linewidth}{!}{
\begin{tabular}{|c|c|c|cc|}
\hline
\textbf{Team}                                           & \textbf{Subtask 1} & \textbf{Subtask 2} & \multicolumn{2}{c|}{\textbf{Subtask 3}}                  \\ \hline
\textbf{}                                               & \textbf{label 1}   & \textbf{label 1}   & \multicolumn{1}{c|}{\textbf{label 1}} & \textbf{label 3} \\ \hline
\begin{tabular}[c]{@{}c@{}}CNLP-\\ NITS-PP\end{tabular} & 0.616              & 0.572              & \multicolumn{1}{c|}{0.616}            & 0.582            \\ \hline
SCalAR & 0.228 & - & \multicolumn{2}{c|}{-} \\ \hline
\end{tabular}
}
\caption{Results of Teams in the Shared Task}
\label{tab:results}
\end{table}

The highest F-1 score was obtained by team CNLP-NITS-PP, they achieved a score of 0.616 for subtask 1, 0.572 for subtask 2 and 0.616 and 0.582 for subtask 3. They were also ranked 1 on the leaderboard. Team SCalAR obtained a F-1 score of 0.228 for subtask 1. 

\section{Conclusion \& Future Work}
This paper summarizes the shared task on gendered abuse detection conducted at ICON 2023. The shared task encompassed of three subtasks which were hosted on Kaggle. We received registration from 9 teams and 2 teams submitted their systems. The winning team, CNLP-NITS-PP, got an F-1 score of  0.616 for subtask 1, 0.572 for subtask 2 and, 0.616 and 0.582 for subtask 3. The dataset is open and will help further the research in abuse detection for Indic Languages. This shared task stands as a meaningful contribution to the broader initiative aimed at fostering a safer online environment. Through building automated approaches and creation of datasets, the task addresses the need to mitigate online gender-based violence, advancing ongoing efforts to enhance internet safety for all.

\section*{Acknowledgment}
We thank the ICON 2023 Conference Committee and the Program Chairs and the Tutorial Chairs in particular for their guidance and support in conducting the shared task. We also express our gratitude to the participating teams for their work.

\bibliography{custom}

\begin{thebibliography}{24}
\expandafter\ifx\csname natexlab\endcsname\relax\def\natexlab#1{#1}\fi

\bibitem[{Arora et~al.(2023)Arora, Jinadoss, Arora, George, Khan, Rawat, Mathur, Yadav, Shora, Raut et~al.}]{arora2023uli}
Arnav Arora, Maha Jinadoss, Cheshta Arora, Denny George, Haseena~Dawood Khan, Kirti Rawat, Seema Mathur, Shivani Yadav, Shehla~Rashid Shora, Rie Raut, et~al. 2023.
\newblock The uli dataset: An exercise in experience led annotation of ogbv.
\newblock \emph{arXiv preprint arXiv:2311.09086}.

\bibitem[{Bhattacharya et~al.(2020)Bhattacharya, Singh, Kumar, Bansal, Bhagat, Dawer, Lahiri, and Ojha}]{bhattacharya2020developing}
Shiladitya Bhattacharya, Siddharth Singh, Ritesh Kumar, Akanksha Bansal, Akash Bhagat, Yogesh Dawer, Bornini Lahiri, and Atul~Kr Ojha. 2020.
\newblock Developing a multilingual annotated corpus of misogyny and aggression.
\newblock \emph{arXiv preprint arXiv:2003.07428}.

\bibitem[{Byman(2021)}]{byman2021hateful}
Daniel~L Byman. 2021.
\newblock How hateful rhetoric connects to real-world violence.

\bibitem[{Chakravarthi et~al.(2021)Chakravarthi, Priyadharshini, Jose, Mandl, Kumaresan, Ponnusamy, Hariharan, McCrae, Sherly et~al.}]{chakravarthi2021findings}
Bharathi~Raja Chakravarthi, Ruba Priyadharshini, Navya Jose, Thomas Mandl, Prasanna~Kumar Kumaresan, Rahul Ponnusamy, RL~Hariharan, John~Philip McCrae, Elizabeth Sherly, et~al. 2021.
\newblock Findings of the shared task on offensive language identification in tamil, malayalam, and kannada.
\newblock In \emph{Proceedings of the first workshop on speech and language technologies for Dravidian languages}, pages 133--145.

\bibitem[{Dadvar et~al.(2013)Dadvar, Trieschnigg, Ordelman, and De~Jong}]{dadvar2013improving}
Maral Dadvar, Dolf Trieschnigg, Roeland Ordelman, and Franciska De~Jong. 2013.
\newblock Improving cyberbullying detection with user context.
\newblock In \emph{Advances in Information Retrieval: 35th European Conference on IR Research, ECIR 2013, Moscow, Russia, March 24-27, 2013. Proceedings 35}, pages 693--696. Springer.

\bibitem[{Frenda et~al.(2019)Frenda, Ghanem, Montes-y G{\'o}mez, and Rosso}]{frenda2019online}
Simona Frenda, Bilal Ghanem, Manuel Montes-y G{\'o}mez, and Paolo Rosso. 2019.
\newblock Online hate speech against women: Automatic identification of misogyny and sexism on twitter.
\newblock \emph{Journal of intelligent \& fuzzy systems}, 36(5):4743--4752.

\bibitem[{Gupta et~al.(2022)Gupta, Roychowdhury, Das, Banerjee, Saha, Mathew, Mukherjee et~al.}]{gupta2022multilingual}
Vikram Gupta, Sumegh Roychowdhury, Mithun Das, Somnath Banerjee, Punyajoy Saha, Binny Mathew, Animesh Mukherjee, et~al. 2022.
\newblock Multilingual abusive comment detection at scale for indic languages.
\newblock \emph{Advances in Neural Information Processing Systems}, 35:26176--26191.

\bibitem[{Kumar et~al.(2020)Kumar, Ojha, Malmasi, and Zampieri}]{kumar2020evaluating}
Ritesh Kumar, Atul~Kr Ojha, Shervin Malmasi, and Marcos Zampieri. 2020.
\newblock Evaluating aggression identification in social media.
\newblock In \emph{Proceedings of the second workshop on trolling, aggression and cyberbullying}, pages 1--5.

\bibitem[{Kumar et~al.(2018{\natexlab{a}})Kumar, Ojha, Zampieri, and Malmasi}]{kumar2018proceedings}
Ritesh Kumar, Atul~Kr Ojha, Marcos Zampieri, and Shervin Malmasi. 2018{\natexlab{a}}.
\newblock Proceedings of the first workshop on trolling, aggression and cyberbullying (trac-2018).
\newblock In \emph{Proceedings of the First Workshop on Trolling, Aggression and Cyberbullying (TRAC-2018)}.

\bibitem[{Kumar et~al.(2021)Kumar, Ratan, Singh, Nandi, Devi, Bhagat, Dawer, Lahiri, and Bansal}]{kumar2021comma}
Ritesh Kumar, Shyam Ratan, Siddharth Singh, Enakshi Nandi, Laishram~Niranjana Devi, Akash Bhagat, Yogesh Dawer, Bornini Lahiri, and Akanksha Bansal. 2021.
\newblock Comma@ icon: Multilingual gender biased and communal language identification task at icon-2021.
\newblock In \emph{Proceedings of the 18th International Conference on Natural Language Processing: Shared Task on Multilingual Gender Biased and Communal Language Identification}, pages 1--12.

\bibitem[{Kumar et~al.(2018{\natexlab{b}})Kumar, Reganti, Bhatia, and Maheshwari}]{kumar2018aggression}
Ritesh Kumar, Aishwarya~N Reganti, Akshit Bhatia, and Tushar Maheshwari. 2018{\natexlab{b}}.
\newblock Aggression-annotated corpus of hindi-english code-mixed data.
\newblock \emph{arXiv preprint arXiv:1803.09402}.

\bibitem[{Kumar et~al.(2014)Kumar, Spezzano, and Subrahmanian}]{kumar2014accurately}
Srijan Kumar, Francesca Spezzano, and VS~Subrahmanian. 2014.
\newblock Accurately detecting trolls in slashdot zoo via decluttering.
\newblock In \emph{2014 IEEE/ACM International Conference on Advances in Social Networks Analysis and Mining (ASONAM 2014)}, pages 188--195. IEEE.

\bibitem[{Mandl et~al.(2020)Mandl, Modha, Kumar~M, and Chakravarthi}]{mandl2020overview}
Thomas Mandl, Sandip Modha, Anand Kumar~M, and Bharathi~Raja Chakravarthi. 2020.
\newblock Overview of the hasoc track at fire 2020: Hate speech and offensive language identification in tamil, malayalam, hindi, english and german.
\newblock In \emph{Proceedings of the 12th Annual Meeting of the Forum for Information Retrieval Evaluation}, pages 29--32.

\bibitem[{Mandl et~al.(2019)Mandl, Modha, Majumder, Patel, Dave, Mandlia, and Patel}]{mandl2019overview}
Thomas Mandl, Sandip Modha, Prasenjit Majumder, Daksh Patel, Mohana Dave, Chintak Mandlia, and Aditya Patel. 2019.
\newblock Overview of the hasoc track at fire 2019: Hate speech and offensive content identification in indo-european languages.
\newblock In \emph{Proceedings of the 11th annual meeting of the Forum for Information Retrieval Evaluation}, pages 14--17.

\bibitem[{Mandl et~al.(2021)Mandl, Modha, Shahi, Madhu, Satapara, Majumder, Sch{\"a}fer, Ranasinghe, Zampieri, Nandini et~al.}]{mandl2021overview}
Thomas Mandl, Sandip Modha, Gautam~Kishore Shahi, Hiren Madhu, Shrey Satapara, Prasenjit Majumder, Johannes Sch{\"a}fer, Tharindu Ranasinghe, Marcos Zampieri, Durgesh Nandini, et~al. 2021.
\newblock Overview of the hasoc subtrack at fire 2021: Hate speech and offensive content identification in english and indo-aryan languages.
\newblock \emph{arXiv preprint arXiv:2112.09301}.

\bibitem[{Mirchandani(2018)}]{mirchandani2018digital}
Maya Mirchandani. 2018.
\newblock Digital hatred, real violence: Majoritarian radicalisation and social media in india.
\newblock \emph{ORF Occasional Paper}, 167:1--30.

\bibitem[{Modha et~al.(2021)Modha, Mandl, Shahi, Madhu, Satapara, Ranasinghe, and Zampieri}]{modha2021overview}
Sandip Modha, Thomas Mandl, Gautam~Kishore Shahi, Hiren Madhu, Shrey Satapara, Tharindu Ranasinghe, and Marcos Zampieri. 2021.
\newblock Overview of the hasoc subtrack at fire 2021: Hate speech and offensive content identification in english and indo-aryan languages and conversational hate speech.
\newblock In \emph{Proceedings of the 13th Annual Meeting of the Forum for Information Retrieval Evaluation}, pages 1--3.

\bibitem[{Mojica(2016)}]{mojica2016modeling}
Luis~Gerardo Mojica. 2016.
\newblock Modeling trolling in social media conversations.
\newblock \emph{arXiv preprint arXiv:1612.05310}.

\bibitem[{Romim et~al.(2021)Romim, Ahmed, Talukder, and Saiful~Islam}]{romim2021hate}
Nauros Romim, Mosahed Ahmed, Hriteshwar Talukder, and Md~Saiful~Islam. 2021.
\newblock Hate speech detection in the bengali language: A dataset and its baseline evaluation.
\newblock In \emph{Proceedings of International Joint Conference on Advances in Computational Intelligence: IJCACI 2020}, pages 457--468. Springer.

\bibitem[{Saroj and Pal(2020)}]{saroj2020indian}
Anita Saroj and Sukomal Pal. 2020.
\newblock An indian language social media collection for hate and offensive speech.
\newblock In \emph{Proceedings of the Workshop on Resources and Techniques for User and Author Profiling in Abusive Language}, pages 2--8.

\bibitem[{Velankar et~al.(2021)Velankar, Patil, Gore, Salunke, and Joshi}]{velankar2021hate}
Abhishek Velankar, Hrushikesh Patil, Amol Gore, Shubham Salunke, and Raviraj Joshi. 2021.
\newblock Hate and offensive speech detection in hindi and marathi.
\newblock \emph{arXiv preprint arXiv:2110.12200}.

\bibitem[{Waseem et~al.(2017)Waseem, Davidson, Warmsley, and Weber}]{waseem2017understanding}
Zeerak Waseem, Thomas Davidson, Dana Warmsley, and Ingmar Weber. 2017.
\newblock Understanding abuse: A typology of abusive language detection subtasks.
\newblock \emph{arXiv preprint arXiv:1705.09899}.

\bibitem[{Zampieri et~al.(2019{\natexlab{a}})Zampieri, Malmasi, Nakov, Rosenthal, Farra, and Kumar}]{zampieri2019predicting}
Marcos Zampieri, Shervin Malmasi, Preslav Nakov, Sara Rosenthal, Noura Farra, and Ritesh Kumar. 2019{\natexlab{a}}.
\newblock Predicting the type and target of offensive posts in social media.
\newblock \emph{arXiv preprint arXiv:1902.09666}.

\bibitem[{Zampieri et~al.(2019{\natexlab{b}})Zampieri, Malmasi, Nakov, Rosenthal, Farra, and Kumar}]{zampieri2019semeval}
Marcos Zampieri, Shervin Malmasi, Preslav Nakov, Sara Rosenthal, Noura Farra, and Ritesh Kumar. 2019{\natexlab{b}}.
\newblock Semeval-2019 task 6: Identifying and categorizing offensive language in social media (offenseval).
\newblock \emph{arXiv preprint arXiv:1903.08983}.

\end{thebibliography}
\bibliographystyle{acl_natbib}
\end{document}